\documentclass[twocolumn,longbibliography, superscriptaddress, english, prb]{revtex4-1}
\usepackage[colorlinks=true,urlcolor=blue,citecolor=blue,linkcolor=blue]{hyperref}
\usepackage[T1]{fontenc}
\usepackage[latin9]{inputenc}
\usepackage{amssymb}
\usepackage{graphicx}
\usepackage{amsmath,color}
\usepackage{mathrsfs}
\usepackage{float}
\usepackage{indentfirst}
\usepackage{braket}
\usepackage{comment}
\usepackage{txfonts}
\usepackage{booktabs}

\usepackage{algpseudocode} 
\usepackage{algorithm}

\def\journal #1, #2, #3, 1#4#5#6{{\sl #1~}{\bf #2}, #3 (1#4#5#6) }

\newcommand{\m}{\mathcal{T}}

\newcommand{\x}{\mathbf{x}}

\def\eqa{\begin{eqnarray}}
\def\eea{\end{eqnarray}}
\newcommand{\eq}{\begin{equation}}
\newcommand{\ee}{\end{equation}}

\renewcommand{\mathbf}[1]{\ensuremath{\boldsymbol{ #1}} }

\usepackage{babel}

\begin{document}
\title{Tree Tensor Networks for Generative Modeling}
\author{Song Cheng}
\email{physichengsong@iphy.ac.cn}
\affiliation{Institute of Physics, Chinese Academy of Sciences, Beijing 100190, China}
\affiliation{University of Chinese Academy of Sciences, Beijing, 100049, China}
\author{Lei Wang}
\email{wanglei@iphy.ac.cn}
\affiliation{Institute of Physics, Chinese Academy of Sciences, Beijing 100190, China}
\author{Tao Xiang}
\email{txiang@iphy.ac.cn}
\affiliation{Institute of Physics, Chinese Academy of Sciences, Beijing 100190, China}
\author{Pan Zhang}
\email{panzhang@itp.ac.cn}
\affiliation{CAS key laboratory of theoretical physics, Institute of Theoretical Physics, Chinese Academy of Sciences, Beijing 100190, China}

\begin{abstract}

Matrix product states (MPS), a tensor network designed for one-dimensional quantum systems, has been recently proposed for generative modeling of natural data (such as images) in terms of ``Born machine''. However, the exponential decay of correlation in MPS restricts its representation power heavily for modeling complex data such as natural images. 
In this work, we push forward the effort of applying tensor networks to machine learning by employing the \textit{Tree Tensor Network} (TTN) which exhibits balanced performance in expressibility and efficient training and sampling. 
We design the tree tensor network to utilize the 2-dimensional prior of the natural images and develop sweeping learning and sampling algorithms which can be efficiently implemented utilizing Graphical Processing Units (GPU).
We apply our model to random binary patterns and the binary MNIST datasets of handwritten digits. We show that TTN is superior to MPS for generative modeling in keeping correlation of pixels in natural images, as well as giving better log-likelihood scores in standard datasets of handwritten digits. 
We also compare its performance with state-of-the-art generative models such as the Variational AutoEncoders, Restricted Boltzmann machines, and PixelCNN. Finally, we discuss the future development of Tensor Network States in machine learning problems. 
\end{abstract}
\maketitle

\section{Introduction}

Generative modeling\cite{statistics-010814-020120}, which asks to learn a joint probability distribution from training data and generate new samples according to it, is a central problem in unsupervised learning.
Compared with the discriminative modeling, which only captures the conditional probability of data's discriminative labels, generative modeling attempts to capture whole joint probability of the data, and is therefore much more difficult\cite{lecun_deep_2015}.

During last decades, there have been many generative models proposed, including those based on Probabilistic Graphic Model (PGM), such as
Bayesian Network~\cite{pearl1985bayesian}, Hidden Markov Model~\cite{huang1990hidden},  and the Restricted Boltzmann Machines~\cite{LeRoux:2008ex}; and models based on neural networks such as the Deep Belief Networks~\cite{hinton2009deep},
the Variational AutoEncoders (VAE)~\cite{2013arXiv1312.6114K}, realNVP~\cite{2016arXiv160508803D},  PixelCNN~\cite{2016arXiv160605328V, 2016arXiv160106759V}, and the recently very popular Generative Adversarial Networks (GAN)~\cite{NIPS2014_5423}.
Among these generative models, there are two models are motivated by physics. One is the Boltzmann machine~\cite{hinton1986learning} where the joint distribution is represented by Boltzmann distribution; and the other one is the Born machine where the Born's rule in quantum physics is borrowed for representing the joint probability distribution of data with squared amplitude of a wave function~\cite{MI_localRBM, han2018unsupervised, 2018arXiv180404168L,li2018shortcut}, and the wave function is represented by tensor networks.

Tensor Networks (TN) are originally designed for efficiently representing quantum many body wave function,~\cite{Orus2014, orus_advances_2014} which, in general, is described by a high order tensor with exponential parameters. TN applies low-rank decompositions to the general tensor by discarding the vast majority of unrelated long-range information to break the so called "exponential wall" of quantum manybody computation. Popular TNs include Matrix Product States (MPS)~\cite{MPSrepresentaion}, Tree Tensor Networks (TTN)~\cite{TTN_Shi}, Multi-scale Entanglement Renormalization Ansatz (MERA)~\cite{MERA}, projected entanglement pair states (PEPS)~\cite{peps}, etc.

In recent years, researchers began to notice the similarities between the tensor networks and the PGM.~\cite{1704.01552, 2013arXiv1301.3124B} Specifically, the factor graph in PGM can be seen as a special kind of tensor  networks\cite{gao_efficient_2017}. In addition to the structural similarities, the problems faced by the TN and PGM are also similar. They both try to use few parameters to approximate probability distribution of an exponentially large number parameters. The reason TN can achieve this is attributed to the physical system's locality.
That is, most of the entanglement entropy of the quantum states we care about obeys the area law~\cite{Eisert:2010hq}. On the PGM side, although the success of machine learning models based PGM in natural images is not completely understood, some arguments support that natural images actually only have sparse long-range correlations, making them much less complex than the arbitrary images~\cite{CheapLearning, MI_localRBM}. Thus the physicist-favor quantum states and the natural images may both only gather in a tiny corner of their possible space, and the similarity between TN and ML models may essentially result from the similarity of the intrinsic structure of the model and data.
Building upon this similarity, various of works have emerged in recent years that applying concept~\cite{Mehta:2014ua, 1704.01552, 2017arXiv171010248P, 2013arXiv1301.3124B}, structure~\cite{2017arXiv170104844D, 2017arXiv170105039G, PhysRevB.97.085104}, and algorithm~\cite{MPSSL, han2018unsupervised, Liu2017} of the tensor networks to machine learning.

In this work, we focus on generative modeling based on tensor networks. On the one hand, we propose TTN as a direct extension to the tree-structure factor graph models; 
On the other hand TTN works as a new tensor network generative model, an extension to the recently proposed MPS Born machine~\cite{han2018unsupervised}. As compared to MPS, TTN exhibit naturally modeling on two dimensional data such as natural images, and it's more favorable in the growth of correlation length of pixels.

In this paper we first introduce TTN as a generative model, then develop efficient sweeping algorithm to learn the model from data by minimizing the Kullback-Leibler divergence between empirical data distribution and the model distribution, as well as a sampling algorithm that generates unbiased samples .
We apply our TTN generative model to two kinds of data. The first one is random binary patterns, where the TTN model works as an associative memory trying to remember all the given patterns. The task is to test the expressive power of the TTN model. The second data we test is the MNIST dataset of handwritten digits, a standard dataset in machine learning.
Using extensive numerical experiments, we show that the TTN has better performance than the classic tree-structure factor graph and the MPS Born machine. In addition, we demonstrate quantitatively the gap between the existing tensor network generation models and the state-of-art machine learning generative models, pointing out the possible future development of the tensor network generation model.


The rest of the paper is organized as follows, in Section \ref{sec:TTN} we give a detailed description of the TTN model, a two-dimensional structure construction, and the training and generating algorithms.  In Section \ref{sec:experiment} we apply the TTN model to both the binary random patterns and the standard binary MNIST dataset. Finally, we discussed the future of tensor network applied to unsupervised generative learning in Section \ref{sec:discussion}.

\section{Models and Algorithms\label{sec:TTN}}
\subsection{The data distribution and maximum likelihood learning}\label{sec:loss}
Suppose we are given a set of data composed of $|\m|$ binary images, $\{\mathbf{x}_a | a= 1,2,3,...,|\m|\}\in \{+1,-1\}^{|\m|\times n}$, each of which is represented by a binary vector of length $n$. This defines a empirical data distribution.
 \begin{align}
 \pi(\mathbf x)=\frac{1}{|\m|}\sum_{a=1}^{|\m|}\delta(\mathbf{x},\mathbf{x_a})
 \nonumber
\end{align}
The task of generative modeling is to find an efficient way to model $\pi\{\mathbf{x}\}$, this means to find a distribution $p(\mathbf{x})$ (with a reasonable number of parameters) which is as close as possible to $\pi\{\mathbf{x}\}$. 
The distance between those two probability can be define by using the Kullback-Leibler (KL) divergence\cite{kullback1951} 
\begin{align}
D_{\textrm{KL}}(\pi\| p)=\sum_{\x} \pi(\mathbf{x})\ln\left(\frac{\pi(\mathbf{x})}{p(\mathbf{x})}\right).
\nonumber
\end{align}
We hence introduce the negative log-likelihood (NLL) as the cost function for model learning:
\begin{equation}
\mathcal{L} = -\frac{1}{|\m|}\sum_{\mathbf{x}\in \mathrm{\m}} \ln\left[{p(\mathbf{x})}\right]=S(p)+D_{\textrm{KL}}(\pi\| p)
\label{eq:nll}
\end{equation}
Where $\m$ indicates the set of given data, and $|\m|$ is the number of training images. Due to the non-negativity of the KL divergence, the last equation indicates that the NLL is bounded below by the Shannon entropy of the dataset. Moreover, since the Shannon entropy $\sum\pi(\mathbf{x})\ln\pi(\mathbf{x})$ is independent of models, minimizing the NLL is equivalent to minimizing the KL divergence. 
\begin{figure}
\begin{center}
\includegraphics[width=\columnwidth]{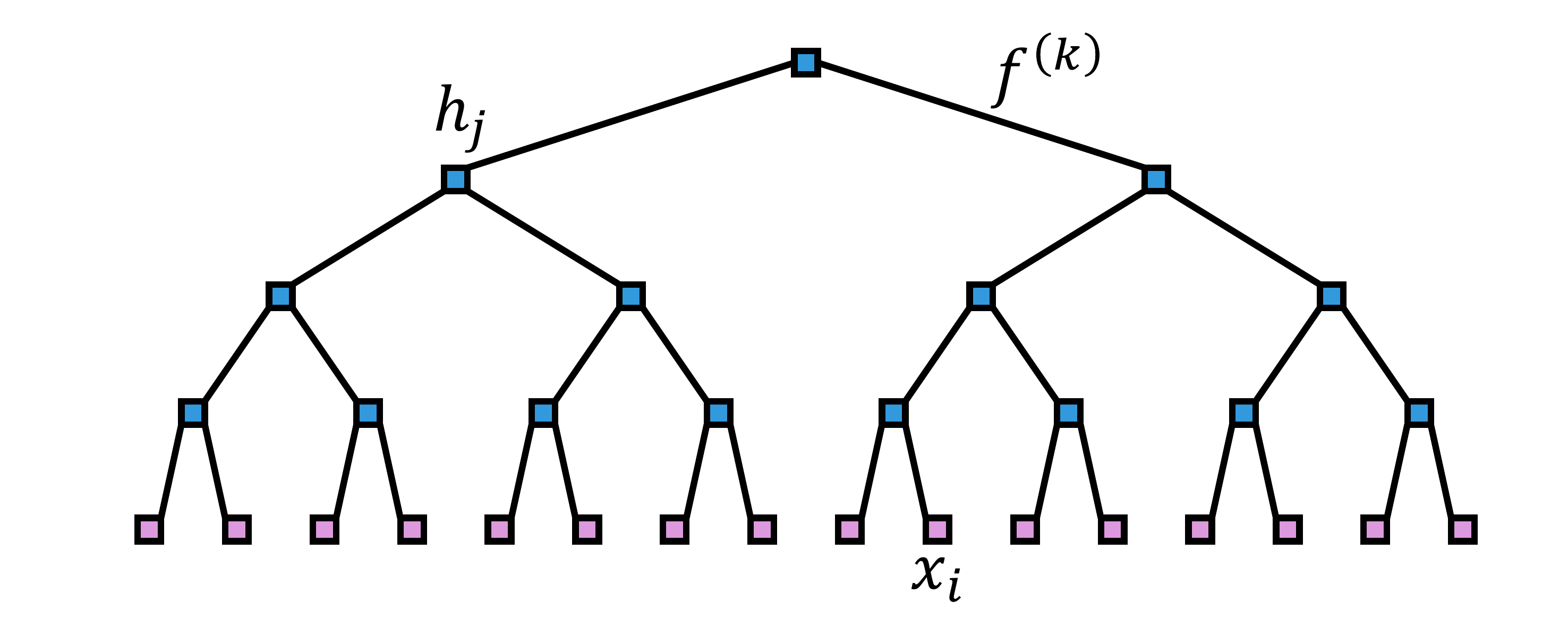}
\large{(a)}
\end{center}
\begin{center}
\includegraphics[width=\columnwidth]{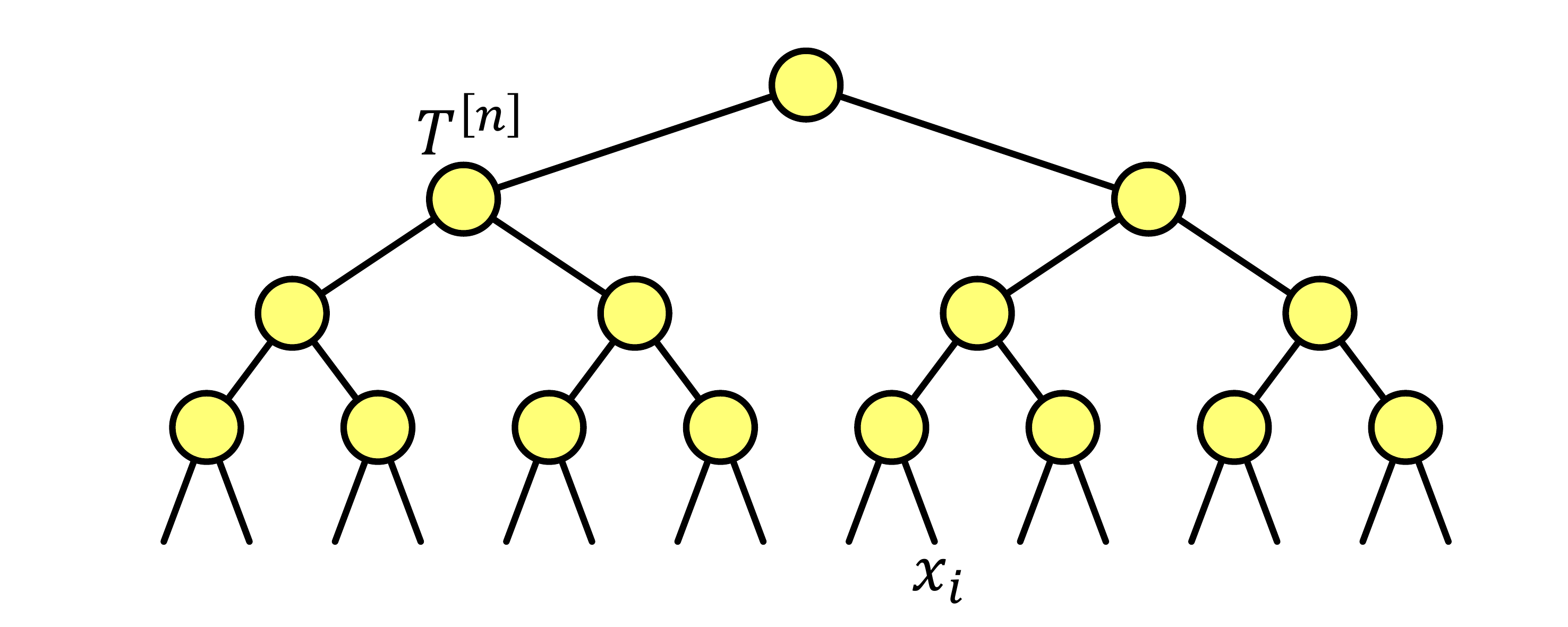}
\large{(b)}
\caption{(a) tree-structure factor graph, where each block denotes a random variable with value -1 or 1, in which the blue/purple block represents a hidden/visible variable respectively. The edge between two blocks introduces a factor function $f^{(k)}$~of those two variables. By adjusting those factor functions, the model could obtain the appropriate joint probability. (b) Tree Tensor Network, where $x_i$ denote the value of the $i$~th pixel of the dataset. Each yellow circle denotes a two or three order tensor. The edge between two tensors denotes a share index of tensors, which is also called "virtue index" in literature and will be contracted later. The exposed edge denote the so called "physical index" of tensors, those indices would ultimately be determined by the dataset. For one of the given configuration of the physical indices, the probability of the configuration would be proportional to the final scale value of the TTN after contracting all the virtue indices.}
\label{fig:ttn}
\end{center}
\end{figure}

\subsection{tree-structure Factor Graph as a generative model}
The art of generative modeling is deeply related to determining a good architecture when representing the best joint probability $p(\mathbf{x})$, which enhances the generalibility. Considering the difficulty of calculating the normalization factor of loop graph, the loop-free PGM like chain or tree are always a relatively simple starting point. Here we take the tree-structure factor graph as an example. 
The un-normalized joint probability $\tilde{p}(\mathbf{x})$ in a tree-structure factor graph represent as
\begin{align}
P(\mathbf{x}) &= \frac{1}{Z}\tilde{p}(\mathbf{x})\nonumber\\ &=\frac{1}{Z} \sum_{\{h_1,h_2,...,h_{N-1}\}} f^{1}(h_1, h_2)f^{2}(h_2, h_3)...f^{2N-2}(h_{N-1}, x_{N})\nonumber
\end{align} 
As shown in Figure \ref{fig:ttn}(a), each block represents a random variable with two states $\{+1,-1\}$. Each purple node $i$ is called a visible node, whose state  $x_i$ is determined by the value of one pixel of the binary input data. Blue node $h_j$ also has two states but they act as hidden variables, hence are not supposed to be observed directly. Each edge of the graph introduces an arbitrary function $f^k$ which maps the states of two-endpoint nodes into a scalar. By combining the scalar on all factors and summing over all possible states of hidden variables $\mathbf{h}$, one gets the non-normalized probability $\tilde{p}(\mathbf{x})$ for a given configuration of pixels~$\mathbf{x}$.

The learning is processed by using gradient descent to minimize the NLL for the given dataset. By denoting the learnable parameters of the model as $\theta$, the gradients read
\begin{equation}
- \nabla_\theta \mathcal{L} = - \frac{1}{|\mathcal{T}|}\sum_{\mathbf{x} \in \mathcal{T}} \nabla_\theta \ln \tilde{p_{\theta}} (\mathbf{x}) + \nabla_\theta \ln Z.
\label{eq:gd_theta}
\end{equation}
In general the first term in the last equation is relatively straightforward to compute. However computing the second term requires of summing over $2^n$ configurations, hence is difficult to compute in general. 
Fortunately, for acyclic graphs such as the tree-structure factor graph, the second term can be computed exactly by using the sum-product algorithm~\cite{Kschischang2006}, a generic message passing algorithm operates on the factor graphs. It's a simple computational rule that by exchanging the multiplication and summation in Z with a certain order to avoid the exponential problem in the brute force summation. 

 It has been proved that any factor graph can be mapped to a tensor network whereas only a special type of tensor networks has its corresponding factor graphs\cite{gao_efficient_2017}. We take the tree-structure graph model as an example. Let us put a matrix $M^{(k)}$ in each edge $k$ and an identity tensor $\delta^{(j)}$ in each hidden node $h_j$, with elements being written as
\begin{equation}
M_{h_a,h_b}^{(k)} = f^k(h_a, h_b), 
\end{equation}
and \begin{equation}
\delta_{l,r,u}^{(j)} = 
\begin{cases}
1, &l=r=u \\
0, &\textrm{otherwise},
\end{cases}
\end{equation} 
where each index of $\delta^{(j)}$ corresponds to an adjacent edge of $h_j$, and bond dimensions of those indices are identical to the number of states of $h_j$.
One can use either QR decomposition or  Singular Value Decomposition (SVD) decomposition to separate the $M^{(k)}$ into a product of two matrices, as
\begin{equation}
M_{h_a,h_b}^{(k)} = \sum_k A_{h_a,k}^{(k)} B_{k, h_b}^{(k)}. 
\end{equation}
Without loss of generality, here we assume in the graph the~$h_a\ge h_b$. The obtained matrices $A, B$ can be absorbed into a tensor defined on nodes, 
\begin{equation}
T_{l,r,u}^{(j)} = \sum_{x,y,z}B_{l,x}^{(l)} B_{r,y}^{(r)} \delta^{(j)}_{x,y,z} A_{z,u}^{(u)}.
\end{equation}
For $j=1$, we simply let the bond dimension of $z, u$ equal to 1. Now we arrive at a specific form of TNN as shown in Fig.~\ref{fig:ttn}(b). Noticed that the tensor $T^{(j)}$ here is just a special subset of the general 3-order tensor, which means if we use general tensors as the building blocks of the TNN, we would get an extension to the origin factor graph model. 

Here we want to remind that the rule of the sum-product approach in tree-structure factor graph is, in fact, equivalent to the tensor contraction of the TTN, with the same order that the sum-product algorithm applies. However, notice that the tensor contraction are much more general than the sum-product algorithm. In those cases that the sum-product algorithm is no longer applicable, the TN can still be  approximately contracted using approaches such as the Tensor Renormalization Group(TRG)~\cite{TRGLevin}.
\subsection{Tree Tensor Network Generative Model}

As motivated in the last section, we treat the TTN as a direct extension of the tree-structure factor graph for generative modeling. 
As illustrated in Figure \ref{fig:ttn}(b), each circle represents a tensor; each edge of the circle represents an individual index of the tensor. The first tensor is a matrix connecting the $2$-nd and $3$-rd tensors. while the remaining tensors are all three-order tensors with three indices. The index between two tensors is called a virtual bond, which would be contracted hereafter. The left and right indices of the tensors in the bottom of the TTN are respectively connected to two pixels of the input image, hence is called physical bonds.

As we have motivated in the introduction, TTN generative model can also be treated as one kind of Born machine~\cite{MI_localRBM}, that is, the TTN represents a pure quantum state $\Psi(\mathbf{x})$, and the $p(\mathbf{x})$ is induced from the square of the amplitude of the wavefunction following Born's rule
\begin{equation}
p(\mathbf{x}) = \frac{|\Psi(\mathbf{x})|^2}{Z}
\end{equation} 
where $Z = \sum_{\mathbf{x}} |\Psi(\mathbf{x})|^2$ is the normalization factor.
In TTN, the $\Psi(\mathbf{x})$ is represented as contraction of totally $N_t$ tensors in TTN,
\begin{equation}
\Psi(\mathbf{x}) = \sum_{\{\alpha\}} T^{[1]}_{\alpha_2, \alpha_{3}}\prod_{n=2}^{N_t}T^{[n]}_{\alpha_{n}, \alpha_{2n}, \alpha_{2n+1}}.
\end{equation}

The reason we choose the quantum inspired Born machine instead of directly modeling a joint probability is based on a belief that the Born machine representation is more expressive than classical probability functions.\cite{2017arXiv170105039G, MI_localRBM} Meanwhile, treating TN as a quantum state could introduce the canonical form of TN, which simplifies the TN contraction calculation and makes contractions more precise. 
For example, if tensor $T^{[2]}$ fulfill $\sum_{\alpha_4, \alpha_5} T^{[2]}_{\alpha_2, \alpha_4, \alpha_5} T^{[2]}_{\alpha_{2}^{'}, \alpha_4, \alpha_5 } = \delta_{\alpha_{2}, \alpha_{2}^{'}}$, we say that the tensor $T^{[2]}$ is canonical for index $\alpha_2$, or more visually speaking, upper-canonical. In the TTN, there are three kinds of canonical forms for each tensor --- the upper-canonical, the left-canonical and the right-canonical respectively, depending on which index was finally left. The three canonical forms are shown in the diagrammatic notation below:
\setlength{\abovedisplayskip}{1pt} 
\begin{equation}
\begin{gathered}
\includegraphics[width=200pt]{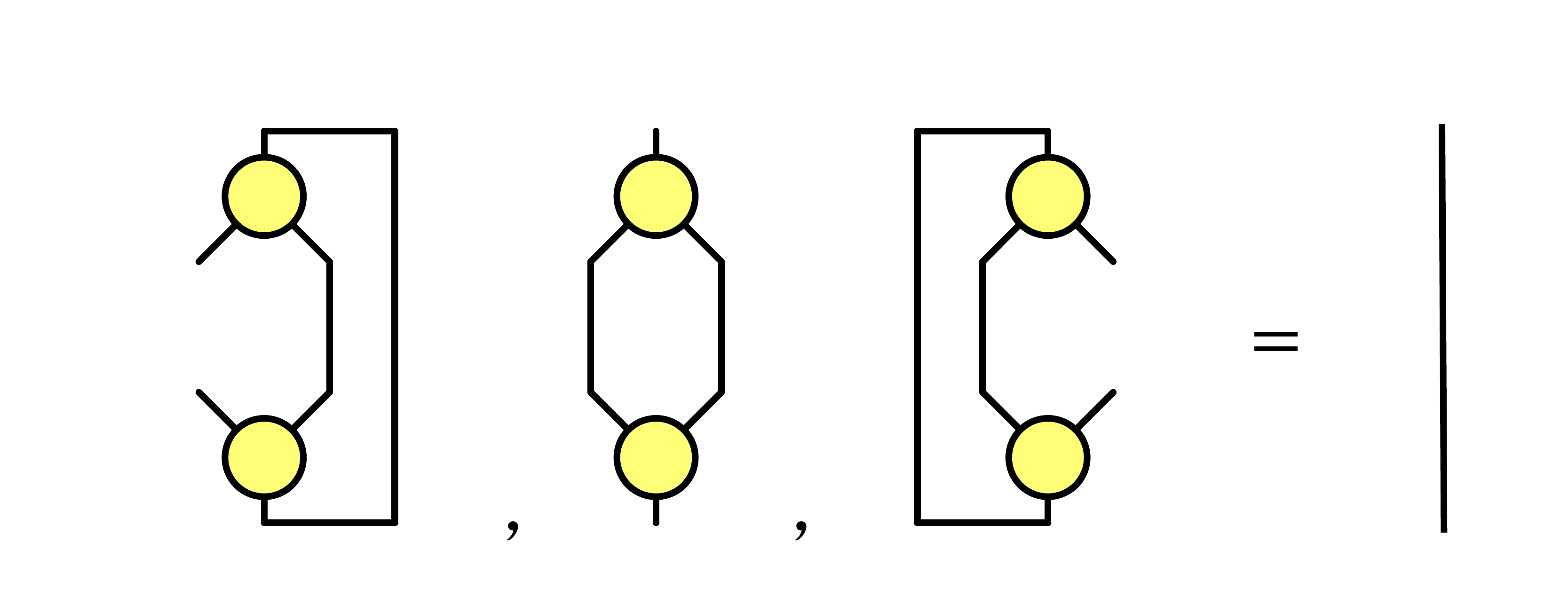} \nonumber
\end{gathered}
\label{eq:canonical}
\end{equation}
the line on the right side represents the identity matrix.

It is technically easy to canonicalize tensor in the TTN. For example, we can start from one end of the tree and use the QR decomposition of the tensor to push the non-canonical part of the tensor to the adjacent tensor. By repeating this step, finally one will push all non-canonical part of the TTN to just one tensor, called the central tensor, and all other tensors are in one of the three canonical forms. Analogous to the mixed canonical form of MPS, we call this form the mixed canonical form of the TTN.

Once the TTN is in the canonical form, many calculations become simple, for example, the normalization factor~$Z$ finally becomes squared norm of a tensor:
\begin{equation}
\begin{gathered}
\includegraphics[width=210pt]{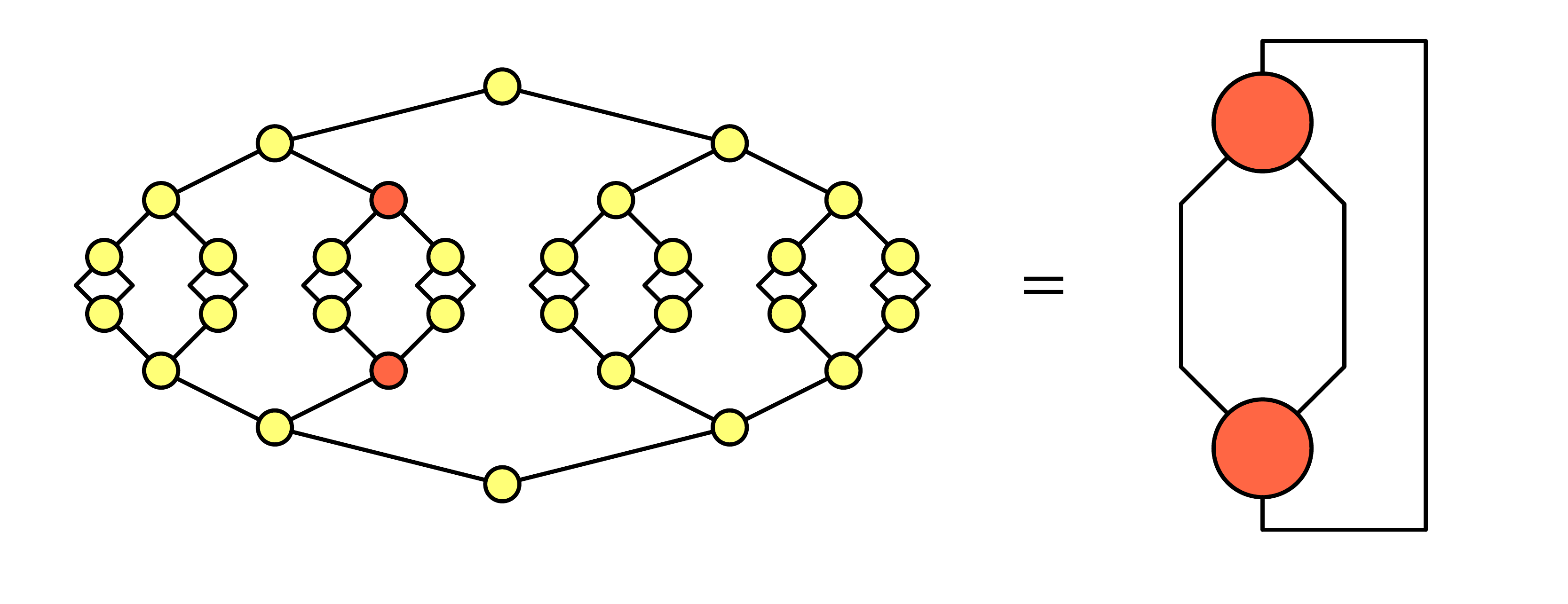},
\label{fig:partition}
\end{gathered}
\end{equation}
where the orange tensor represents the non-canonical central tensor, on an arbitrary position. The direction of all tensor's canonical form is pointed to the direction of the central tensor. 
After all, to get the normalization $Z$, the only calculation we need to do is the trace of multiplication of the central tensor by its complex conjugate. 

General tensor networks have gauge degree of freedom on its virtual bond. One can insert a pair of unitary matrices $UU^{-1}$ in the virtual bond without changing the final contraction results. This could damage the accuracy of the training algorithm and brings additional computational complexity. Fortunately, for acyclic tensor networks like the TTN, the canonical form fixes this degree of freedom.

\subsection{Data representations}
\begin{figure}
\begin{center}
\includegraphics[width=\columnwidth]{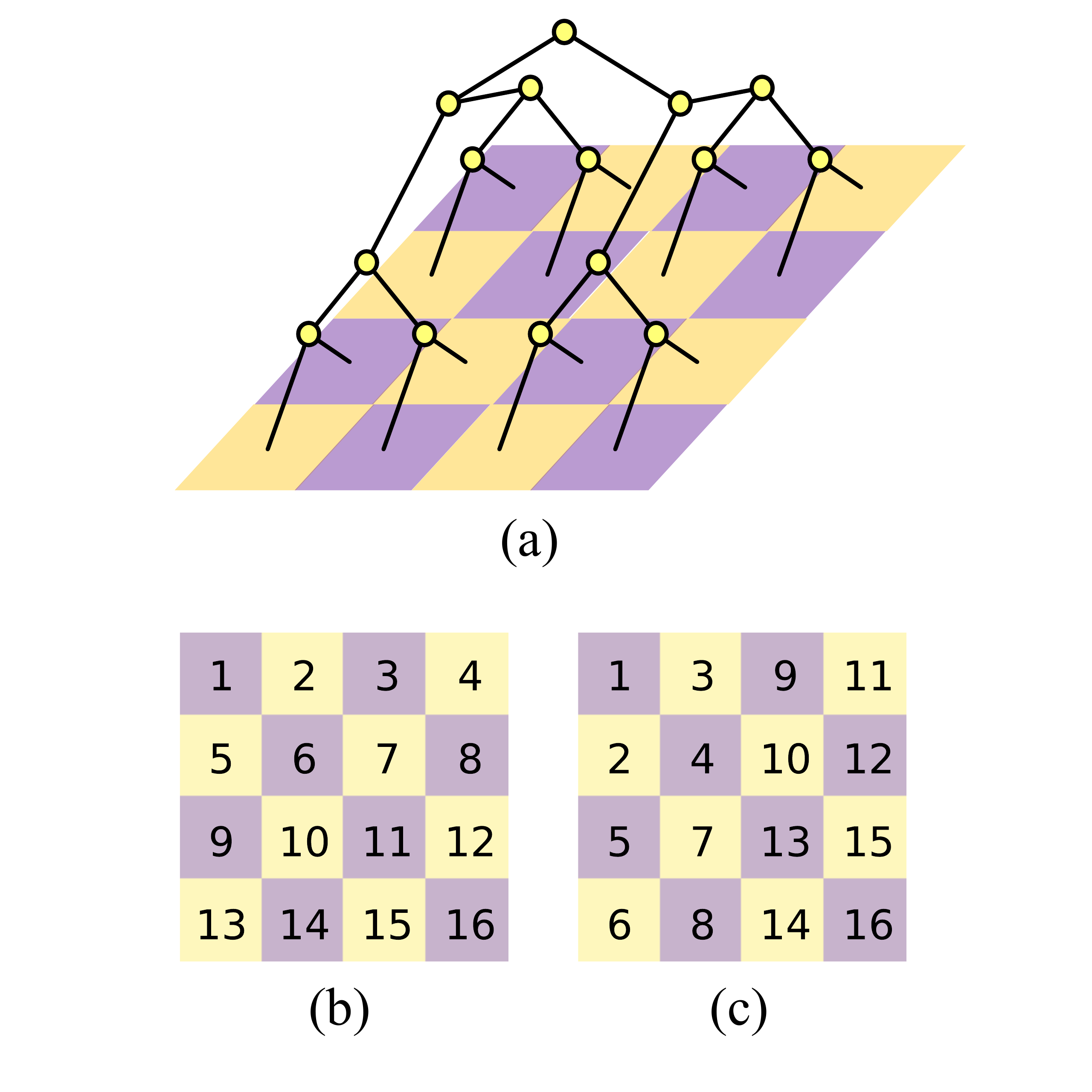}
\caption{(a), the TTN with 2d structure. Changing the 1d order of data with the 2d order is equivalent to use the TTN with 2d structure replacing Figure \ref{fig:ttn}(b); (b), the 1d order of data; (c), the 2d order of data. } 
\label{fig:1d2d}
\end{center}
\end{figure}
In this work, we consider binary data, such as black and white images, so the local dimension of the Hilbert space of each physical bond is $2$. As illustrated in Figure~\ref{fig:1d2d}, each index for the lowest layer tensors have two components, corresponding to the two possible values of the connected pixels. The pixels can be simply vectorized from the image to a vector, as explored in \cite{han2018unsupervised} for the MPS Born machine, which we call \textit{ $1$-D representation}, as it basically does not use any features in the $2$-D structure of the images.

Compared with the MPS, there is a significant advantage of the TTN, that it can easily achieve the two-dimensional modeling of natural images. Figure \ref{fig:1d2d}(a) shows the two-dimensional modeling of TNN. In this architecture, each tensor is responsible for one local area of pixels, which greatly reduces the artificial fake long-range correlations. Hence we call it \textit{$2$-D representation}. Clearly, the $2$-D representation keeps the model structure of Figure \ref{fig:ttn}, while only requires reshuffling the data index to proper order, as shown in Figure \ref{fig:1d2d}(b)(c). \cite{2007PhRvL, 2009PhRvB}

In practice, in order to ensure that the number of input pixels is a power of 2, we may artificially add some pixels that are always zero. If the input data is the 1d permutation, we add those zero pixels to the two ends of the one-dimensional chain; if it is 2d, we add to the outermost of the 2d lattice. This is analogous to ``padding'' operation in convolution networks.

\subsection{Training algorithm of the TTN}
As we introduced in Sec.~\ref{sec:loss}, the cost function we used in the training is the Negative Log Likelihood Eq.\eqref{eq:nll}, which is also the KL divergence between the target empirical data distribution and the probability distribution of our model, up to a constant.

To minimize the cost function, a standard way is the Stochastic Gradient Descent algorithm (SGD). Unlike traditional SGD, which updates all trainable parameters at the same time, in the TTN we have a sweeping process, that is, iteratively updates each tensor based on the gradient of the cost function with respect to tensor elements of a tensor while holding other tensors unchanged. This sweeping process can be combined with the canonicalization form of the tensor network to simplify computations. As formulated in Eqs.~\eqref{fig:partition} and \eqref{eq:partial}, after canonicalization, the whole network is equivalent to one single tensor, which significantly improves the efficiency of the sweeping process. 
There are two choices of the updating scheme: single-site update scheme in which we update a single $3$-way tensor at one time with other tensors hold; and two-site update scheme in which we first merge two neighboring tensors then update the merged $4$-way tensors with other tensors hold.
For the single-site update, the gradient reads
\begin{equation}
\frac{\partial \mathcal{L} }{\partial T^{[k]}} = \frac{Z'}{Z} - \frac{2}{|\mathcal{T}|}\sum_{\mathbf{x} \in \mathcal{T}}\frac{\Psi'(\mathbf{x})}{\Psi(\mathbf{x})}
\label{eq:gd}
\end{equation}
where $\Psi'(\mathbf{x})$ and $Z'$ denotes the derivative of $\Psi(\mathbf{x})$ and $Z$ with respect to the $T^{[k]}$, they are depicted by the following diagram:
\begin{equation}
\begin{gathered}
\includegraphics[width=200pt]{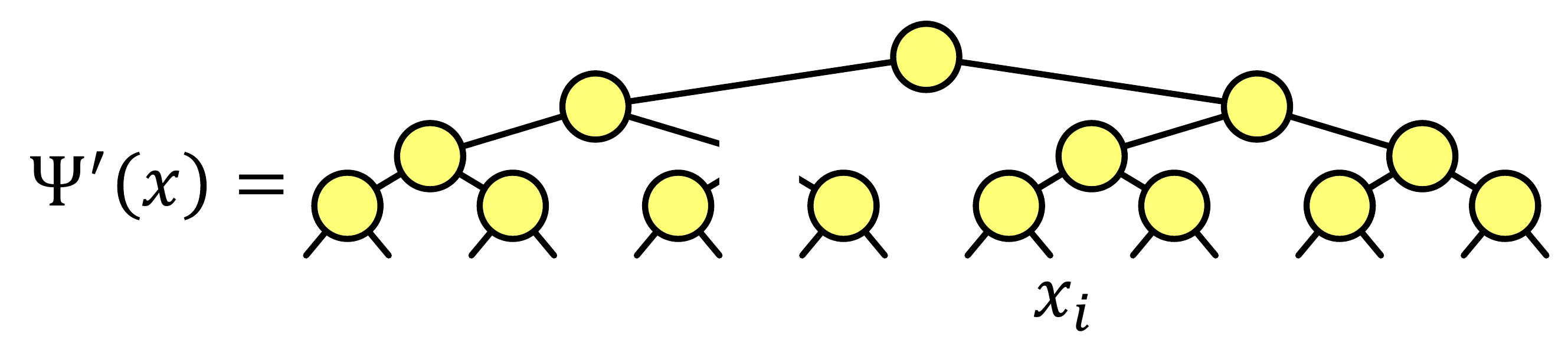} \\
\includegraphics[width=170pt]{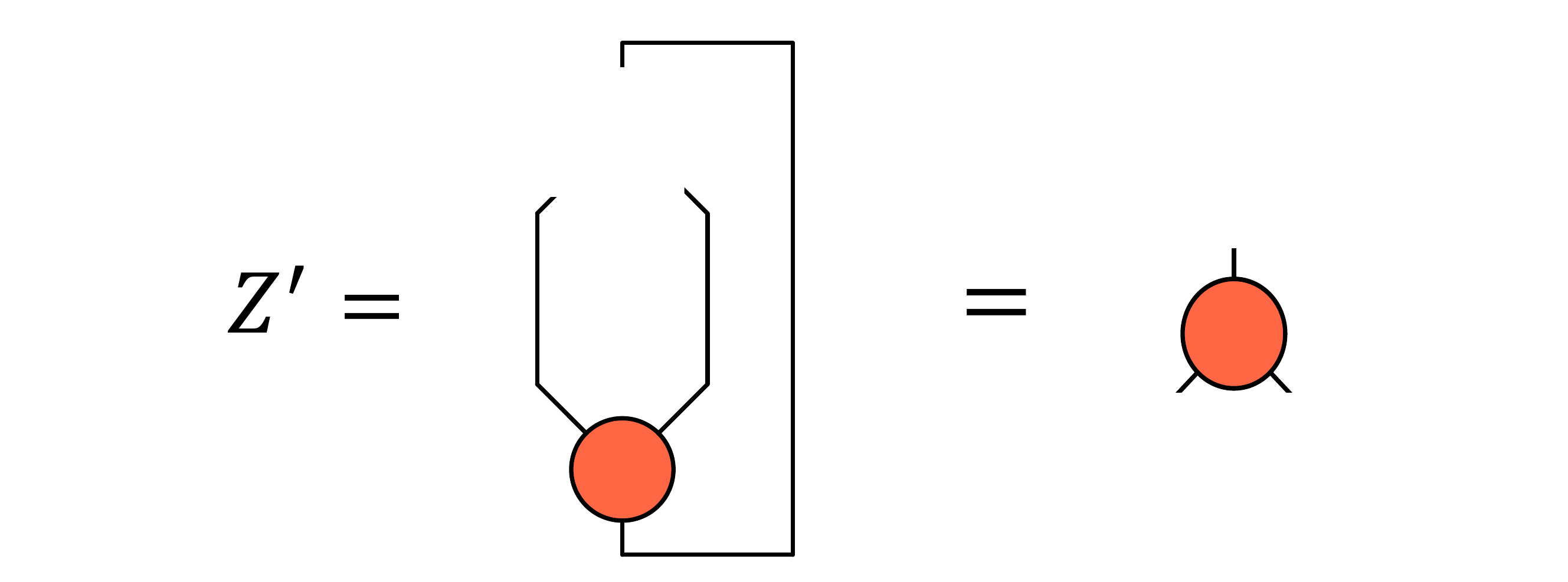} \\
\label{eq:partial}
\end{gathered}
\end{equation}

As already noted, thanks to the tree canonicalization, computation of $Z$ become straightforward. 

The first step of the training is to transform the TTN consisting of many random initialized tensors into the canonical form, then push the non-canonical part onto the first tensor to be trained, e.g. the rightmost one. Next, we use the gradient calculated by Eqs. \eqref{eq:gd}, \eqref{eq:partial} to update corresponding tensors. 
\begin{equation}
T^{[k]}_{new} = T^{[k]} - \alpha \frac{\partial \mathcal{L}}{\partial T^{[k]}},
\label{eq:update}
\end{equation}
where $\alpha$ denotes the learning rate; then we move to the next sweeping step. For maintaining the canonical form, as shown in Algorithm~\ref{alg:sweep}, each time we apply a QR decomposition to the updated tensor, store the orthogonal matrix $Q$ as a new tensor and contract $R$ to the tensor which is going to be updated in the next step.

If we start from the rightmost tensor of the TTN, this rule will allow us to gradually traverse the entire TTN from right to left. Then we chose the leftmost tensor as our starting tensor, doing the entire update again from left to right. A complete left-right-left sweeping defines an epoch of learning. See Algorithm~\ref{alg:sweep} for the details of the training algorithm.

\begin{algorithm}[H]
	\begin{algorithmic}[1]
		\caption{Sweeping algorithm of the TTN.}
		\Require Tensors $T^{[i]}$ in the TTN. The TTN has been canonicalized towards the rightmost tensor $T^{[N]}$.
		\Ensure  Updated tensor $T^{[i]}_{new}$. The TTN will be canonicalized towards the rightmost tensor $T^{[N]}_{new}$.
		\State Mark all tensors as "unupdated". Set $T^{[N]}$ as the current tensor $T^c$. 
		\While{Exist unupdated tensor}
    		\If{Exist one unupdated adjacent tensor of $T^c$.}
    		    \State Update $T^c$ by the SGD. Mark this tensor as "updated".
    		    \State Set the rightmost unupdated adjacent tensors of $T^c$ as the next $T^c$.  
		        \State Apply QR decomposition on the previous $T^c$. Reshape $Q$ to the shape of the previous $T^c$, save it as $T_{new}$.  Contract $R$ to next $T^c$. 
    		\ElsIf{Exist two unupdated adjacent tensors of $T^c$.}
    		    \State Do 5-6.
    		\EndIf 
    	\EndWhile 
    	\State Mark all tensors as "unupdated".
    	\State Sweep from left to right. 
         \label{alg:sweep}
         \end{algorithmic}
\end{algorithm}

For the two-site update, most of the procedures are the same as the single-site update. The only difference is that the tensor to be updated is a $4$-way tensor  $M^{[k, j]}$ merged by two $3$-way tensors. After using the gradient of $\mathcal{L}$ on the merge tensor to update the merge tensor, we apply SVD on the merge tensor to rebuild the two tensors of size identical to the original two tensors before merging, while pushing the non-canonical part onto the next tensor. 
Each SVD step gives us a chance to change the dimension of the bond between the current tensor and the last tensor, making the TTN support dynamical adjustment of number of parametrers of the model. This is the main benefit of the two-site update scheme compared to the one-site update one. It is also an important advantage of the tensor network compared to traditional machine learning algorithms.

Notice that the one-site update always has lower computational complexity ($~O(D^3)$ than the two-site update $~ O(D^5)$). In our experience, although the one-site update needs more epochs to converge, its final convergence result is not significantly different from the two-site update.
\subsection{Direct sampling of the TTN generative modeling}
\label{sec:sample}
Unlike most of the traditional generative models, the TTN can directly calculate the partition function exactly. This gives TTN an ability to sample configurations directly without needing the Markov chain i.e. Gibbs sampling.
We first compute the marginal probability of arbitrary pixel $k$. 
\begin{equation}
p(x_k) = \frac{\sum_{\mathbf{x_a}, \forall i \neq k}|\Psi(\mathbf{x})|^2}{Z},
\label{eq:marginal}
\end{equation}
where the numerator in graphical notation is quite similar to that of computing $Z$ with the only difference that the bond corresponding to $x_k$ does not contract and $\Psi(\mathbf{x})$ left as a two-dimensional vector. The square of the values of this two-dimensional vector are marginal probability of $x_{k} = 0, 1$ respectively.
Then the conditional probability for the next pixel is computed as
\begin{equation}
p(x_j | x_k) = \frac{p(x_j, x_k)}{p(x_k) }.
\label{eq:condition}
\end{equation}
In diagram notation this is equivalent to using sampled value of $x_k$ to fix the corresponding bond of $x_k$, and keep the corresponding bond of $x_j$ open in contraction.
The conditional probability of Eq.~\eqref{eq:condition} can be generalized to the case of multiple fixed pixels. Equipped with all the conditional probabilities, we are able to sample pixels of images one by one. 


\section{Numerical Experiments\label{sec:experiment}}

\subsection{Random dataset}

\begin{figure}
\begin{center}
\includegraphics[width=\columnwidth]{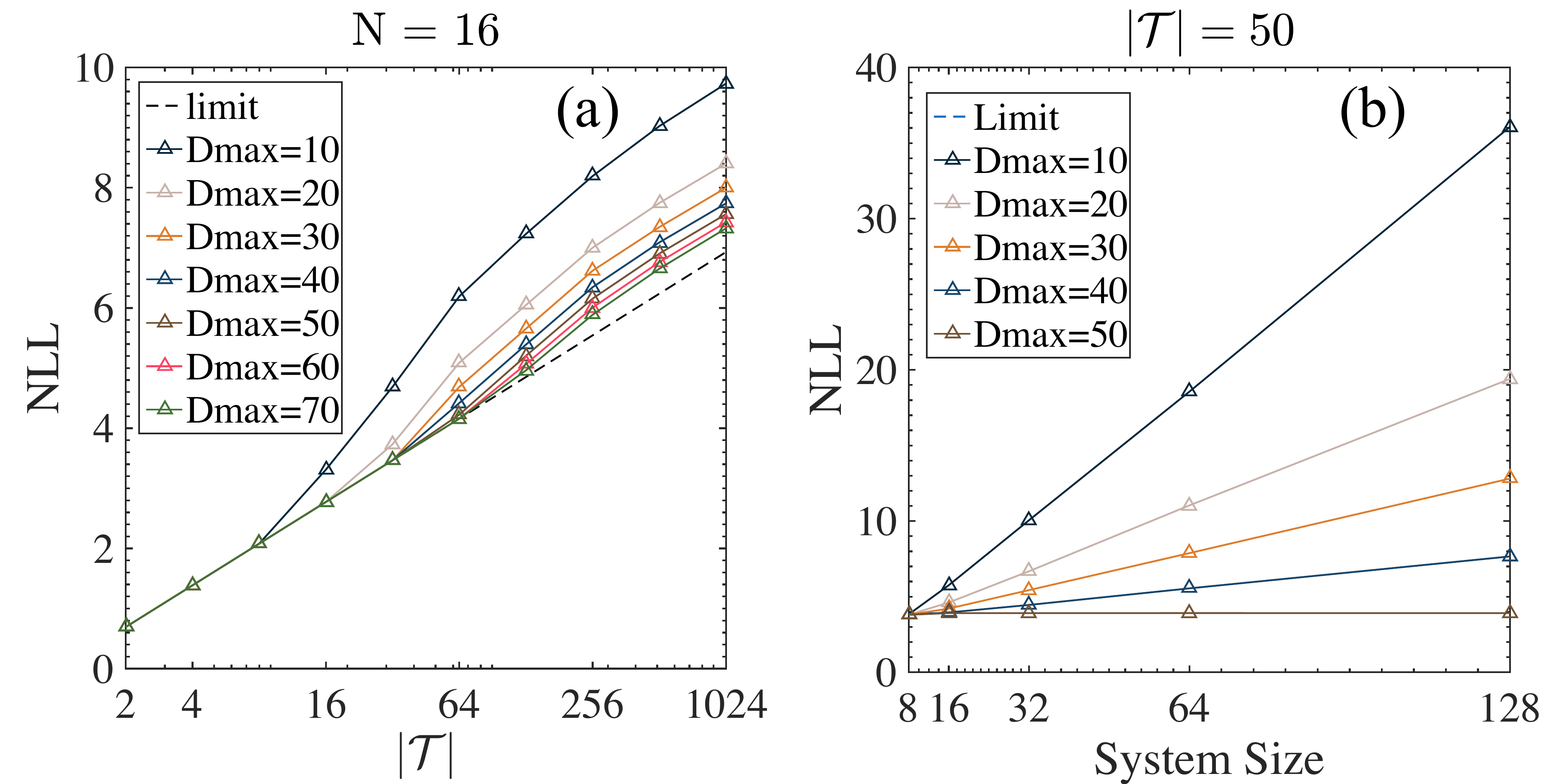}
\caption{(a) Training NLL of TTN Born machine as a function of the data size $|\mathcal{T}|$, the system size is $N=16$. (b) Training NLL of TTN Born machine as a function of the system size $N$, the data size $|\mathcal{T}|=50$.}
\label{fig:random_nll}
\end{center}
\end{figure}

Remembering a specific set of random samples, i.e. as an associative memory \cite{hopfield_neural_1982}, is perhaps the hardest and the least biased task for testing the expressive power of generative models.  
Since in TTN we are able to compute the partition function, nomalized probability of the training sample, as well as NLL exactly, we can quantify how good our model learned from the training random samples. General speaking, the smaller the NLL, the more information that we have captured from the training dataset. Notice that the theoretical lower bound of NLL is $\ln(|\m|)$. Thus if NLL is equal to $\ln(|\m|)$, it means the KL divergence is zero, indicating that the distribution of our model is exactly the same as empirical data distribution. That is, our model has exactly recorded the entire training set, and is able to generate samples identical to training data with an equal probability assigned to each of the training samples.

In Figure \ref{fig:random_nll} (a), we show the NLL of training set as a function of the number of training patterns $|\m|$. The dashed line is the NLL's theoretical limit $\ln(m)$. As we can see, the NLL does not converge to the theoretical limit when the maximum bond dimension $D_{max} < |\m|$. 
The reason for this phenomenon is that, in the traditional theory of tensor networks, the maximum information entropy that a bond of the tensor network can captured equals to $\ln(D)$.

\begin{figure}
\begin{center}
\includegraphics[width=\columnwidth]{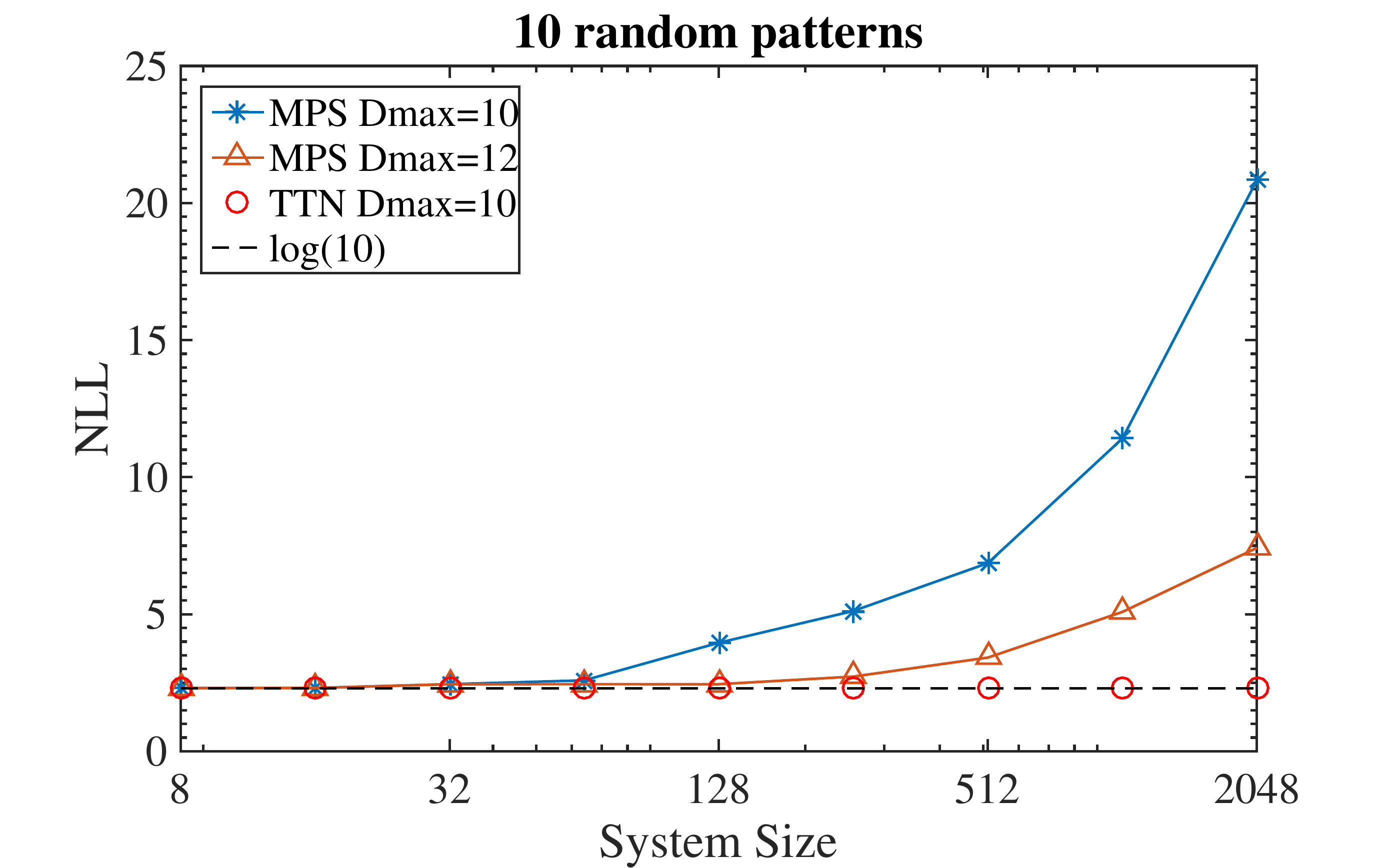}
\caption{Comparison between TTN and MPS Born machines trained on $10$ random patterns with different system sizes. As the system size become larger, MPS can no longer reach the theoretical limit of NLL when  $D_{max}$ equal to the number of samples, while the TTN is almost unaffected by the system size. This is because the structure of the TTN can capture better the long-range dependences.} 
\label{fig:rand}
\end{center}
\end{figure}

In Figure \ref{fig:random_nll} (b) we plot NLL as a function of the number of pixels in each random pattern. The number of training patterns $m = 50$. The figure shows that when $|D_{max}| < |\mathcal{T}|$ the NLL increases almost linearly with the number of variables in the pattern. This is because the long-range correlations of a particular set of random patterns are dense, and the TTN does not have enough capacity to exactly record all the information of the random patterns. 
When $|D_{max}| \ge m$, since the correlation length of pixels in the TTN only logarithmic growth with the size of the image, the NLL can always easily converge to the theoretical limit regardless of how big the size of the picture is.

This point is further illustrated in Fig~\ref{fig:rand} where the relationship between system size and training NLL on different models are compared. As an example, we use $|\m|=10$ random patterns for training both the TTN and the MPS models. We found that even at very large $N$, the TTN can still converge to NLL's theoretical minimum once its maximum bond dimension reaches to $10$. However, under the same or even higher bond dimension ($D_{max} = 12$), NLL of the MPS still fails in converging to the theoretical bound when the size is very large.
Because in the MPS, the correlation length decays exponentially fast, this makes the information contained in the middle bond more saturated when the image size becomes very large, making the maximum likelihood training less efficient.

\subsection{Binary MNIST dataset}
A standard benchmark for computer vision, especially for the generative modeling, is the handwritten digits of the MNIST dataset. The binary MNIST dataset contains a training set of $50,000$ images, a validation set of $10,000$ images, and a test set of $10,000$ magess. Each of them is handwritten digits of $28 \times 28$ pixels of value 0 or 1. 
In order to facilitate comparison with other work, we directly use the same standard binary MNIST dataset that has been used in the analysis of Deep Belief Networks and been widely recognized by the machine learning community~\cite{salakhutdinov2008}. The dataset can be downloaded directly from the corresponding website.\cite{datalink}
 
\begin{figure}
\begin{center}
\includegraphics[width=\columnwidth]{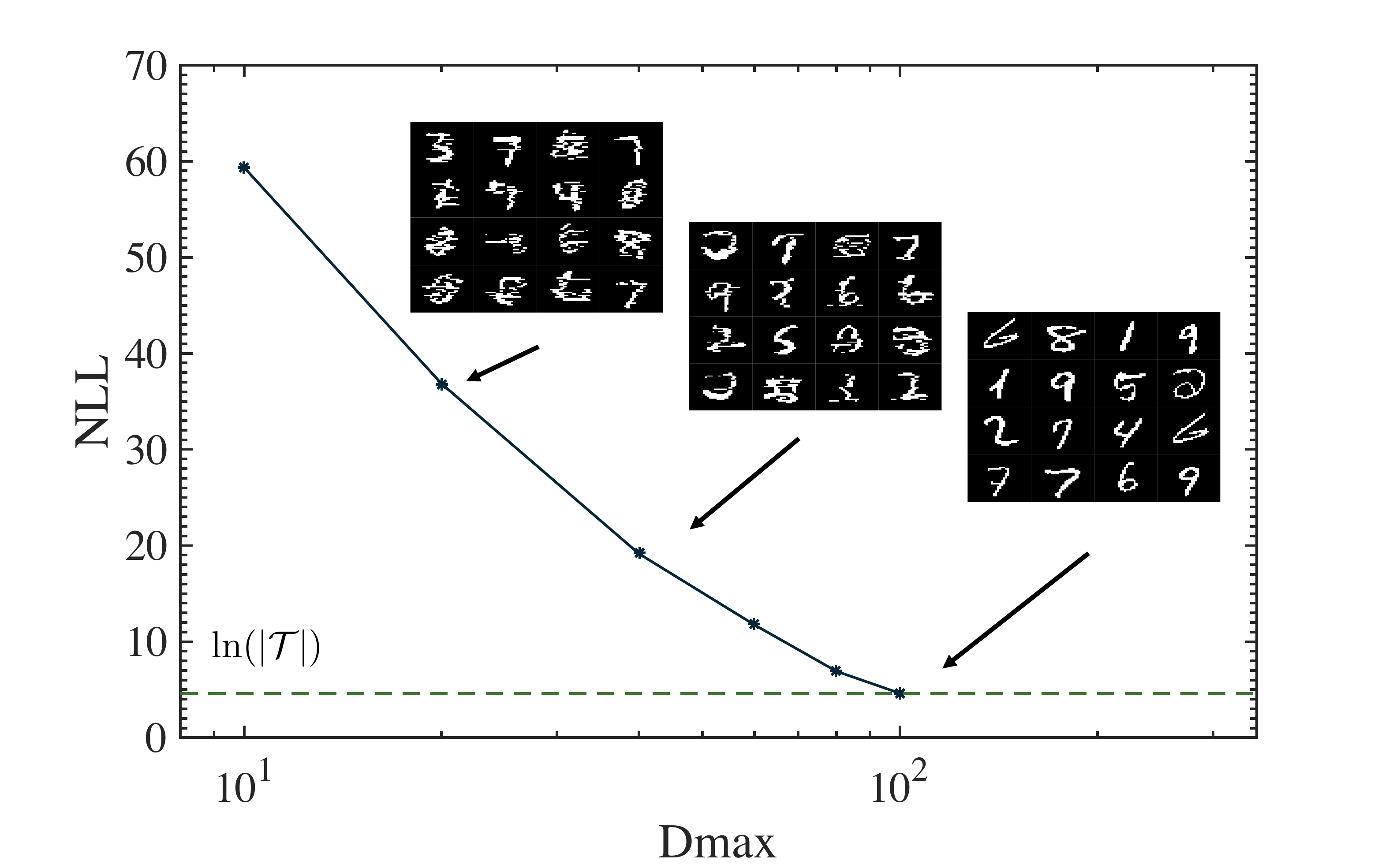}
\caption{Training NLL and sampling images for $|\mathcal{T}| = 100$ binarized MNIST dataset. $\ln(|\mathcal{T}|)$ is the theoretical minimum of NLL. The TTN exactly remembers all the information of the images when $D_{max} = |\mathcal{T}|$.}
\label{fig:sample100}
\end{center}
\end{figure}

We did three experiments on the binary MNIST dataset. In the first experiment we use $100$ randomly selected images for training TTN with different $D_{max}$. The results are shown in Figure \ref{fig:sample100} where we can see that with the NLL gradually decreases, the quality of the generated samples becomes better. The training NLL would decrease to its theoretical minimum as $D_{max}$ increasing to $|\mathcal{T}|$ where the sampling image will be exactly the same as one in the training set.%

\begin{figure}
\begin{center}
\includegraphics[width=250pt]{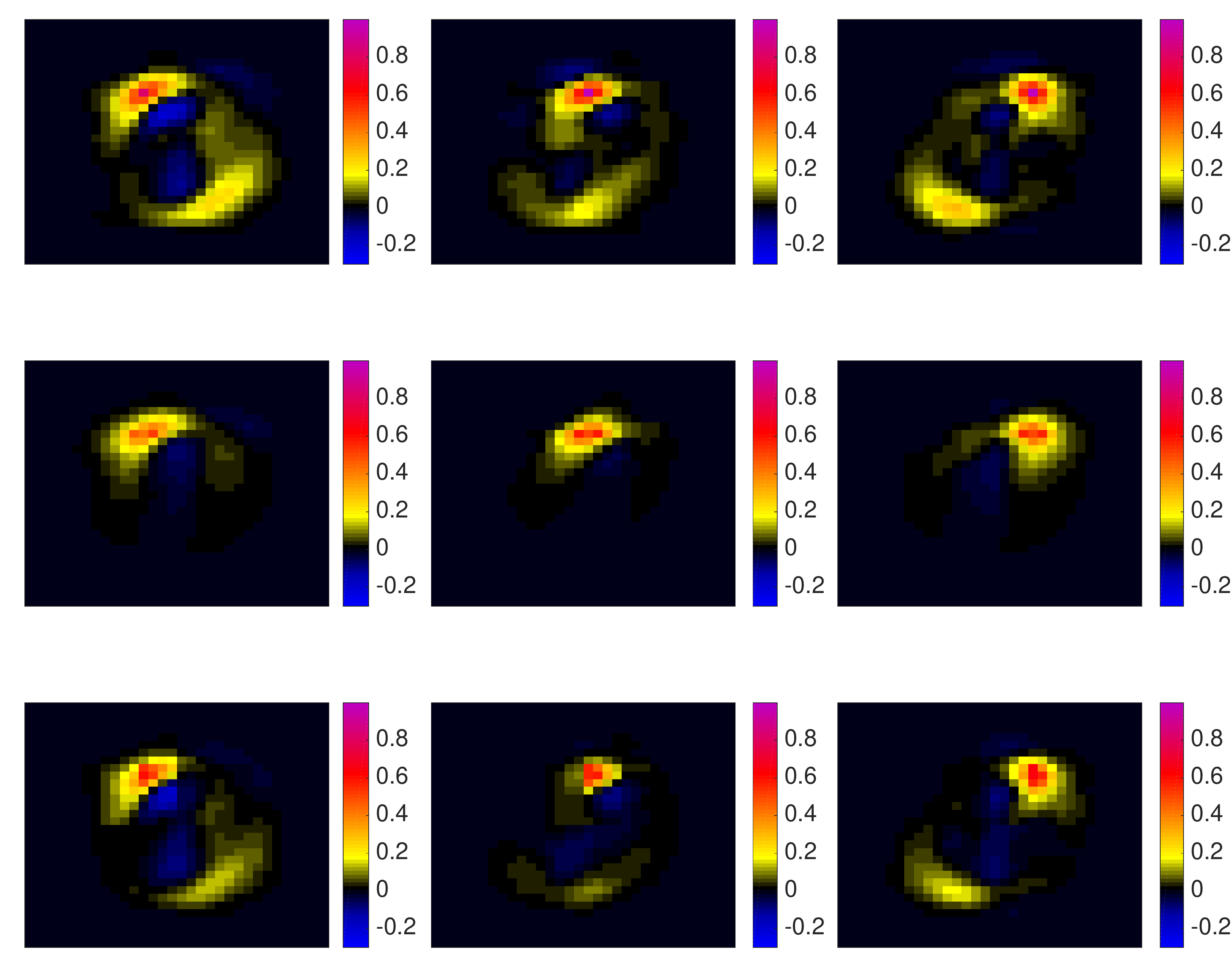}
\caption{Two-site correlation of pixels extracted from the original data (1st row), the MPS (2nd row) and the TTN model (3rd row). 
We randomly choose three pixels at the 10th row of the images. The $D_{max}$ of TTN is 50, the $D_{max}$ of MPS is 100, which correspond to the models with the smallest test NLL. }
\label{fig:correlation}
\end{center}
\end{figure}

In Figure \ref{fig:correlation} we plot the two-site correlation function of pixels. In each row, we randomly select three pixels, then calculate the correlation function of the selected pixels with all others pixels. The values of the correlations are represented by color. The real correlations extracted from the original data is illustrated in the top row, and correlations constructed from learned MPS and TTN are shown in the bottom rows for comparison. For TTN and MPS, the $D_{max}$ is 50 and 100 respectively, which correspond to the models with the smallest test NLL.
As we can see that in the original dataset, the correlation between pixels consists of short-range correlation and a small number of long-range correlation. However, the MPS model can faithfully represent the short-range correlation of pixels, while the TTN model performs well in both short-range and long-range correlations. 


Next we carry out experiments using the whole MNIST dataset with $50,000$ training images to quantitatively compare the performance of TTN with existing popular machine learning models. 
The performance is characterized by evaluating NLL on the $10,000$ test images. 
We also applied the same dataset to the tree-structure factor graph and the MPS generative model, and compare on the same dataset the test NLL with RBM, Variational AutoEncoder (VAE) and PixelCNN which currently gives the state-of-the-art performance. Among these results, RBM and VAE only evaluate approximately the partition function, hence gives only approximate NLL. While TTN, MPS, together with PixelCNN are able to evaluate exactly the partition function and give exact NLL values.

The results are shown in Table~\ref{nll-label}, where we can see that the test NLL obtained by the tree-structure factor graph is 175.8, the result of MPS is 101.45, with corresponding $D_{max} = 100$. While for the TTN on $1$-D data representation (as depicted in Fig.~\ref{fig:1d2d}(b)) with $D_{max} = 50$, the test NLL already reduces to $96.88$. With the same $D_{max}$, the TTN performed on $2$-D data representation (as depicted in Fig.~\ref{fig:1d2d}(a,c)) can do even better, giving NLL around $94.25$.
However, we see from the table that when compared to the state-of-the-art machine learning models, the tensor network models still have a lot of space to improve: the RBM using $500$ hidden neurons and $25$-step contrastive divergence could reach NLL approximately $86.3$, and the PixelCNN with 7 layers gives NLL around $81.3$. 


\begin{figure}
\begin{center}
\includegraphics[width=\columnwidth]{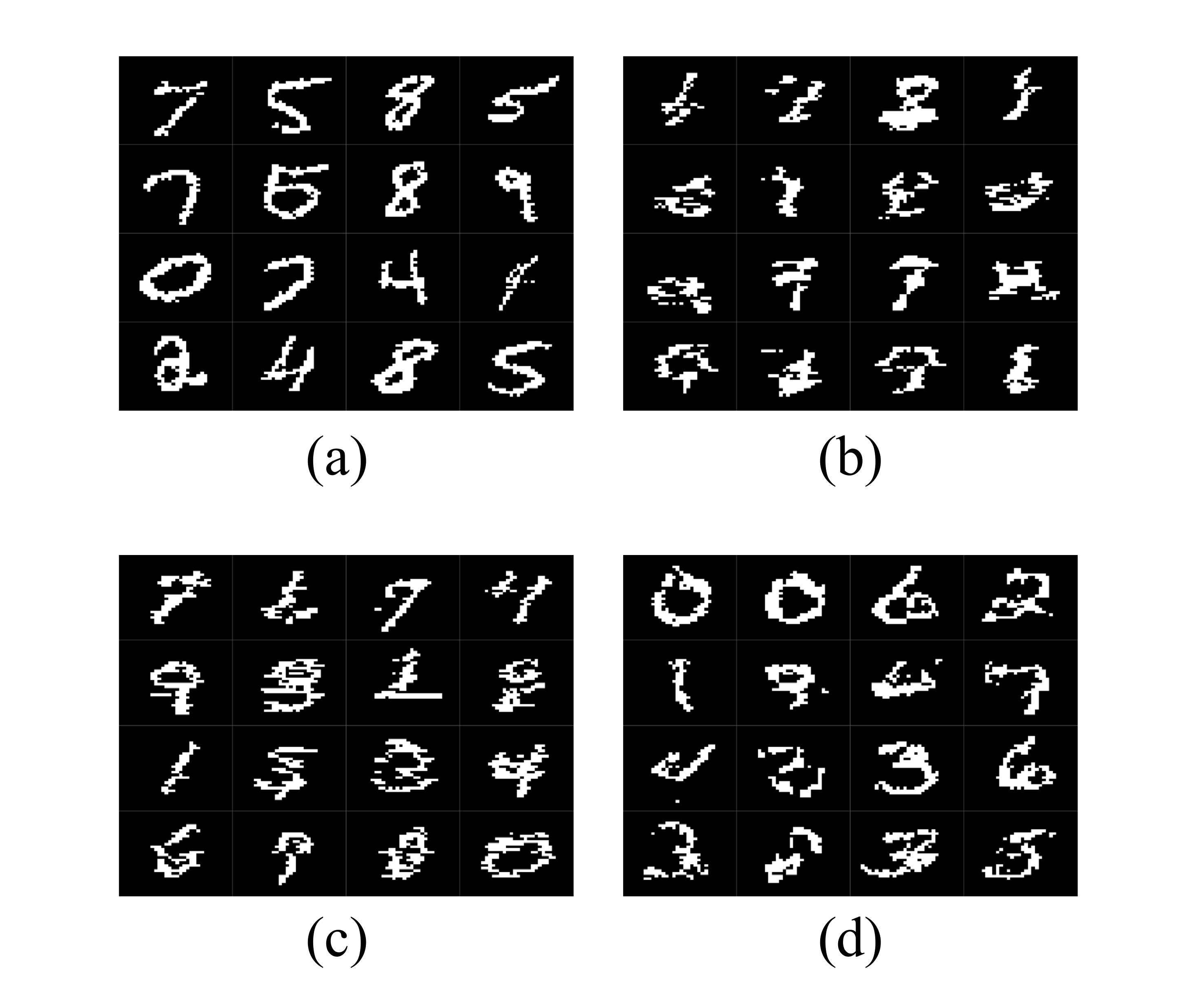}
\caption{Images generated by MPS and TTN that are trained on $|\mathcal{T}| = 50000$ training images. (a): Part of the training images; (b): MPS with $D_{max} = 100$ , test NLL = 101.45; (c): 1d-TTN with $D_{max} = 50$, test NLL = 96.88; (d): 2d-TTN with $D_{max} = 50$, test NLL = 94.25.}
\label{fig:samples}
\end{center}
\end{figure}

\begin{table}
\centering
\caption{Test NLL of different model for binary MNIST dataset}
\label{nll-label}
\begin{tabular}{@{}lllll@{}}
\\
\toprule
Model & \multicolumn{4}{l}{Test NLL} \\ \midrule
Tree factor graph & \multicolumn{4}{c}{175.8}  \\
MPS & \multicolumn{4}{c}{101.5}  \\
TTN-1d  & \multicolumn{4}{c}{96.9}   \\
TTN-2d  & \multicolumn{4}{c}{94.3}   \\
RBM  & \multicolumn{4}{c}{86.3*}~\cite{salakhutdinov2008}   \\
VAE  & \multicolumn{4}{c}{84.8*}~\cite{burda_importance_2015}   \\
PixelCNN & \multicolumn{4}{c}{81.3}~\cite{2016arXiv160106759V}  \\ \bottomrule 
* stands for approximated NLL. 
\end{tabular}
\end{table} 

In Figure \ref{fig:samples} we draw the sampled images from TTN trained on $50,000$ MNIST images, using the sampling algorithm described in Sec.~\ref{sec:sample}, and compare them with the images sampled from MPS trained on the same dataset. It shows that TTN with $2$-D data representation samples are better eye-looking than MPS figures, indicating that TTN captures better global dependences than MPS.

\section{Conclusions and Discussions \label{sec:discussion}}
We have presented a generative model based on the tree tensor networks. This model is a direct extension of the matrix product state Born machine~\cite{han2018unsupervised} and also a generalization the tree-structures factor graph for generative modeling. The TTN inherits advantages of MPS on generative modeling, including tractable normalization factor, the canonical form and the direct sampling, but overcomes the issue of exponential decay of correlations in MPS, 
making it more effective in capturing long-range correlations and performing better in large size images. It is also straightforward to perform TTN on the two-dimensional modeling of images. We have developed efficient sweeping training algorithms for decreasing NLL lost function using single site as well as two site updating schemes.

We have carried out extensive experiments to test the performance of the proposed TTN Born machine and compare it with existing approaches.
We showed that TTN gives better training NLL than MPS (with the same bond dimension) on remembering large random patterns. on classic MNIST handwritten digits, TTN captures long ranger dependences better than MPS, and gives much better NLL on test images, which indicates a better generalization power. 

Naturally, a further development to the current work is by introducing the structure of multi-scale entanglement renormalization ansatz (MERA)\cite{2012PhRvB85p5146F, 2012PhRvB85p5147F}, another type of tensor network we can expect to have tractable partition function while hopefully being able to preserve better the long range dependences in the data.

We have also pointed out the gap between current generative models based on the tensor networks and the state-of-the-art machine learning models based on neural networks such as the PixelCNN. One advance of neural network based models is the better prior for the images powered by the convolution. So an important step for tensor-network based models to improve further is utilizing better priors of $2$-D images.
Along with this direction, the Projected Entanglement Pair States (PEPS)\cite{peps}, which gives much better prior to natural images, should be considered. However, notice that this comes with the compensation that partition function is no longer exactly computable. 
It might be not a serious problem as approximate contraction algorithms such as Tensor RG, boundary MPS, and Corner transfer matrix have been proved to be efficient in contracting PEPS for finite-size systems. We will put this into future work.

We emphasize here that the necessity of developing generative learning algorithms based on tensor network mainly motivated by the quantum machine learning field. The machine learning model based on tensor network representation is essentially the type of model that used a specific quantum state to represent classical data. Research on this type of model will pave the way for future migration of machine learning to quantum computers~\cite{gao_efficient_2017}.

On the other hand, traditional machine learning models dealing with generative learning links closely to tensor networks dealing with quantum many body problems. Finding out the reason machine learning model performs better than the tensor network on generative learning can also help the traditional tensor network algorithms continue to improve.

\begin{acknowledgments}
We thank E. Miles Stoudenmire, Jing Chen, Xun Gao, Cheng Peng, Zhao-Yu Han, Jun Wang for inspiring discussions and collaborations. 
S.C. and T.X are supported by the National R\&D Program of China (Grant No. 2017YFA0302901) and the National Natural Science Foundation of China (Grants No. 11190024 and No. 11474331).
L.W. is supported by the Ministry of Science and Technology of China under the Grant No. 2016YFA0300603 
and National Natural Science Foundation of China under the Grant No. 11774398. P.Z. is supported by Key Research Program of Frontier Sciences of CAS, Grant No. QYZDB-SSW-SYS032 and Project 11747601 of National Natural Science Foundation of China.
\end{acknowledgments}

\bibliography{ttn}

\end{document}